# Automatic Ontology Construction Using LLMs as an External Layer of Memory, Verification, and Planning for Hybrid Intelligent Systems

Pavel Salovsky  
Partenit.io  
San Francisco, CA, USA  
psalovsky@gmail.com

Iuliia Gorshkova  
Partenit.io  
San Francisco, CA, USA  
yu.gorshkova@gmail.com

**Executive summary.** The paper proposes and formalizes a neuro-symbolic architecture in which a large language model is not treated as a self-sufficient knowledge store, but is used as a layer of interpretation, generation, and orchestration over external memory. This memory is represented by an ontological graph in RDF/OWL standards, supplemented by a vector RAG storage, SHACL validators, SPARQL query mechanisms, dialogue logs, and embeddings. The main hypothesis is that an LLM can automatically build and update an ontology from texts, dialogues, and external sources, and then use this graph as a structured, verifiable, and long-term model of a domain. This hypothesis aligns with modern work on LLM-assisted ontology engineering, GraphRAG, and external context orchestration.

In the empirical part, project materials provided by the author are reconstructed and carefully interpreted: architectural diagrams, a Tower of Hanoi benchmark chart, and Fact Analyzer screenshots. The quantitative result on Tower of Hanoi shows that ontological augmentation increases success rate for 3 disks from 26.3% to 33.3%, for 5 disks from 33.3% to 45.5%, with parity at 4 disks and zero results at 6 disks. Since the number of runs, exact prompts, temperature, success criteria, and statistical aggregation method are not disclosed in the source materials, these results are treated as descriptive evidence rather than final statistical proof. Nevertheless, they qualitatively align with literature showing that purely autoregressive LLMs face systematic difficulties in planning tasks and benefit from integration with external models, constraint-checking, and symbolic procedures.

**Abstract**. The article presents a full formulation of the problem of automatic ontology construction using large language models and their application as external memory and a verification layer for hybrid intelligent systems. The proposed architecture combines MCP orchestration, vector RAG storage, an RDF/OWL graph, SPARQL queries, SHACL validation, reasoning engines, dialogue logs, and an agent layer. Special attention is given to the Ontology Builder pipeline: entity extraction, relation extraction, normalization, triple formation, validation, and recording in TTL/RDF. The empirical part is based on provided project materials and includes quantitative analysis of Tower of Hanoi and a qualitative case of strict verification of a regulatory question. It is concluded that ontology not only improves retrieval but shifts the system from a "search overlay" to a structured world model enabling long-term memory, explainability, contradiction control, and more robust planning.

memory; MCP; RDF; OWL; SPARQL; SHACL; reasoning; agents; Tower of Hanoi

## Introduction

Over the past two years, RAG has become the standard engineering response to an obvious limitation of LLMs: the model is capable of formulating, summarizing, and explaining effectively, but it does not reliably store up-to-date knowledge beyond its trained weights and current context window. As demonstrated by the series of articles by P.Salovskii on Habr ([habr.com/ru](habr.com/ru)), even the simplest form of RAG indeed functions as a connection to external memory: documents are split into chunks, encoded as embeddings, retrieved by similarity, and supplied to the model's context. However, as data volume and question complexity increase, it becomes evident that such RAG remains primarily a retrieval overlay rather than a полноценная knowledge system. It retrieves similar fragments, but does not necessarily reconstruct relationships, rules, and causality.

The key intellectual gap can be formulated as follows: when a human reads a text, they rarely store knowledge as a linear sequence of phrases. Instead, the text is continuously transformed into structure—into relationships, roles, causes, effects, temporal hierarchies, and constraints. This idea is central in P.Salovskii's articles *"From Text to Knowledge"* and *"Memory for AI and Robots"*: while an LLM can plausibly imitate structural understanding in its responses, its own memory is not organized as an explicit graph of entities and relationships. This leads to the idea of externalizing structure into a separate layer—an ontology—where knowledge is stored not as an array of texts, but as a network of facts, states, and relationships.

This transition becomes especially important for agent-based and robotic systems. As long as the task is short and local, an LLM can "live" within the current context. However, once multi-session memory, decision history, user profiling, long-term learning processes, environmental evolution, or real-world plan execution are required, the weakness of short-term memory becomes the primary limiting factor. In this context, external ontological memory acts not merely as a knowledge base, but as an operating system for stable state accumulation and verifiable action.

This paper systematizes this idea. The contributions of the work are as follows. First, the project architecture is reconstructed and formalized based on the provided diagrams, where the LLM is connected via MCP to an ontological graph, a vector RAG storage, API tools, logs, and embeddings. Second, a complete pipeline for an automatic Ontology Builder is defined, suitable for both publication and engineering implementation. Third, quantitative and qualitative results are described based on the Tower of Hanoi benchmark and the Fact Analyzer case. Fourth, a broader thesis is derived from recently published scientific and technological sources: ontology can be understood as long-term, structured, and verifiable external memory, while the LLM functions as a generator, a query compiler, and an interface to this memory.

## Related Work

The symbolic foundation of the proposed architecture is defined by the official W3C specifications. RDF defines a graph-based data model as a set of triples of the form subject–predicate–object; OWL provides a formalized vocabulary of terms and relationships and allows an ontology to be treated either as an abstract structure or as an RDF graph; SPARQL formalizes the syntax and semantics of queries over RDF; SHACL defines a constraint language for validating data graphs against a shapes graph. Taken together, this stack provides what purely vector-based retrieval lacks: formal structure, typing, constraints, and verifiability.

Over the past two years, a distinct research direction has emerged focused on applying LLMs to ontology engineering tasks. The survey by Garijo et al. shows that ontologies remain a key component of knowledge engineering for integration, validation, and reasoning, but their development remains labor-intensive and requires manual involvement; at the same time, the authors note the rapid growth of work on generating competency questions, ontology learning, alignment, and validation using LLMs. A more recent survey by Bian systematizes not just individual techniques, but a broader transition from rule-based and statistical pipelines to language-driven and generative pipelines across three levels: ontology engineering, knowledge extraction, and knowledge fusion.

Several specific works are particularly important for our design. Kommineni et al. demonstrated the viability of a semi-automated pipeline in which an LLM formulates competency questions, assists in constructing the TBox, and is subsequently used to populate a knowledge graph, while the final system still requires a human-in-the-loop for quality evaluation. Lippolis et al. investigated not only the feasibility of generating OWL drafts, but also their structural quality, showing that properly designed prompting regimes can produce ontologies comparable to those created by novice ontology engineers, although variability and frequent errors remain. Nayyeri et al., in the RIGOR framework, showed that RAG can be used not only for answers over documents but also for the iterative generation of richer OWL ontologies from relational schemas using a judge-LLM and provenance-tagged delta fragments. Feng et al. further demonstrated ontology-grounded knowledge graph construction under the Wikidata schema as a way to improve interpretability and consistency in automated extraction. Aggarwal et al., using the IEEE Thesaurus, showed that LLMs reliably identify semantic relationships between scientific topics and that well-tuned smaller models can sometimes approach the quality of larger ones at significantly lower cost.

A separate line of work concerns not ontologies per se, but graph-based retrieval. The work by Edge et al. on GraphRAG shows that vanilla RAG performs poorly on "global" questions over large corpora, which require not retrieval of a local fragment but query-focused summarization and sensemaking. In response, GraphRAG constructs an entity graph and community summaries, which are then incorporated into the prompt. The survey by Han et al. establishes this as a general GraphRAG framework, while Microsoft Research documentation emphasizes that knowledge graphs and community summaries are used to augment prompts when querying large corpora. For the purposes of this paper, the key conclusion is that graph representations improve retrieval not merely through embedding similarity, but through the structural organization of the knowledge source.

Finally, closely aligned in system intuition are the context-centric approach of Interfaze and the position regarding the role of small models in agentic AI. Interfaze frames modern LLM applications as the task of "building and acting over context," where a compact structured state is constructed on top of a heterogeneous DNN/SLM stack, and the final large model operates on an already distilled context. Belcak et al. independently demonstrate that for agent-based systems, where individual modules repeatedly solve small specialized tasks, small language models are often more economically and operationally suitable, and heterogeneous systems represent a natural design. This aligns with the formulation proposed here, in which the agent layer handles graph retrieval, reasoning, validation, fact-checking, and action abstraction, while the main LLM remains a high-level generator.

## Methods and Architecture

The article is based on two classes of materials. The first class consists of external sources: official specifications, arXiv preprints, scientific surveys, documentation on GraphRAG, reasoning, and SHACL, as well as three P.Salovskii articles and a public description of Pavel Salovsky's seminar. The second class consists of author-provided project materials: an md summary, two architectural diagrams, a Tower of Hanoi chart, and Fact Analyzer screenshots. To strictly separate verifiable claims from project artifacts, the following rule is applied: general statements and definitions are grounded in external sources; numerical and interface observations derived from project images are described as author-supplied evidence.

The central architectural hypothesis is as follows: the LLM should operate not directly on "raw" documents and not solely through vector similarity, but through a context layer in which structured ontological memory, vector retrieval, and a tool bus coexist. This role is fulfilled in modern tooling ecosystems by MCP: the specification defines it as an open protocol that enables sharing context with language models, exposing tools/capabilities, and constructing composable workflows based on JSON-RPC 2.0 between hosts, clients, and servers. In the proposed system, MCP is used as a universal communication bus between the LLM, ontology graph, vector database, external APIs, and agents.

**Figure 1.** Reconstructed architecture of a hybrid system based on the provided project diagrams. The diagram combines project-specific topology with formally standardized MCP components and context-centric orchestration.

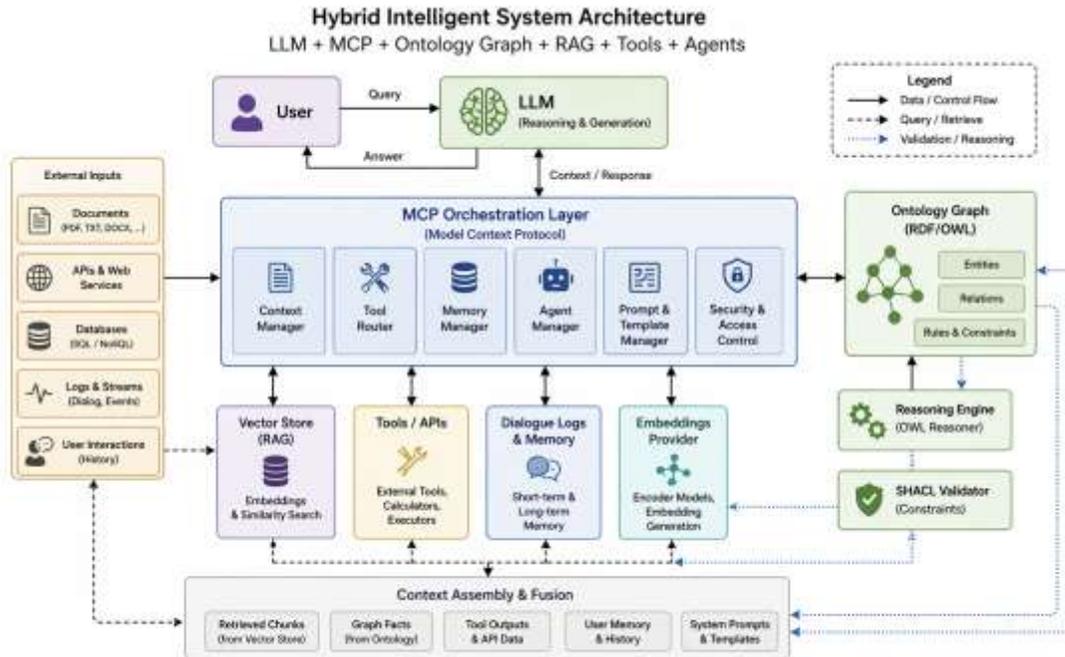

Figure 1. Reconstructed architecture of a hybrid system based on the provided project diagrams. The diagram combines project-specific topology with formally standardized MCP components and context-centric orchestration.

Flowchart description:

User query → LLM (reasoning + planning)
LLM ↔ MCP (orchestrator)
MCP ↔ Vector RAG storage
MCP ↔ RDF/OWL graph
MCP ↔ External APIs and services
MCP ↔ Files, databases, documents
Graph ↔ Reasoning engines
Graph ↔ SHACL validation
Documents → Ontology Builder
APIs → Ontology Builder
Vector DB → Context
Graph → Context
APIs → Context
Documents → Context
Context → MCP
LLM → Final answer
Answer → Fact extraction from response and dialogue
Fact extraction → Ontology Builder
Fact extraction → Logs and embeddings
Logs → Vector DB
Ontology Builder → TTL/RDF artifact

TTL → Graph DB
Graph → Agent layer
Agents → MCP

From the perspective of data representation, the architecture relies on dual memory. The first memory is an RDF/OWL graph G, where entities, relationships, types, and constraints are explicitly stored. The second memory is a vector store V, where embedding representations of source and derived texts are stored. For a query q, a composite context is constructed:

$C(q) = Fuse(R\_vect(q), R\_graph(q), R\_tool(q), M\_user)$,

where R_vect is vector retrieval over chunks and logs, R_graph is SPARQL-based extraction of neighborhoods and rules from the graph, R_tool represents external tool calls via MCP, and M_user is user or session memory. This duality allows RAG and ontology not to be opposed, but functionally separated: the vector store handles recall and textual residue, while the graph provides semantics, constraints, and explainability. This composition aligns with both GraphRAG and modern context-centric systems.

The functional comparison of components is shown below.

| Component | Main function | Type of knowledge | What it validates/enhances |
|---|---|---|---|
| Vector RAG storage | fast recall by similarity | distributed embeddings | relevant text fragments |
| RDF/OWL graph | structured memory | entities, relations, types, axioms | typing, relations, logical consequences |
| SPARQL | graph access | graph patterns, paths, filters | precise graph-aware queries |
| SHACL | graph validation | shape constraints | structural correctness and completeness |
| Reasoning engine | classification and inference | axioms and rules | consistency and implications |
| MCP | orchestration of tools and context | protocols + tool schemas | workflow composition and access to external systems |
| Logs/embeddings | audit trail and auxiliary memory | traces, histories, dense vectors | reproducibility and retracing |

| Component | Main function | Type of knowledge | What it validates/enhances |
|---|---|---|---|
| Agent layer | execution and specialization | plans, tool policies, validators | action loop and multi-step control |

In this formulation, "ontology as memory" ceases to be a metaphor. It means that from dialogue and actions, not only narrative text is extracted, but also changes in the state of the world model: who is related to whom, when, under what conditions, and in what status. This distinguishes a system with cumulative memory from a chat history that must be repeatedly summarized. This interpretation directly aligns with P.Salovskii's argument that long-term knowledge about the user, task, and environment is best represented as a network of entities, events, states, and relationships.

## Ontology Builder Pipeline

At the level of the engineering process, the Ontology Builder represents a closed pipeline that takes documents, dialogues, API responses, and tables as input, and produces an ontology delta in TTL/RDF format as output. Unlike classical ontology engineering, where conceptualization and encoding were often performed manually, here most of the work is delegated to LLMs and accompanied by symbolic validation. The literature shows that such semi- and semi-automated pipelines are becoming the primary practical approach today: from competency questions and OWL drafts to judge-LLM, ontology-grounded extraction, and iterative ontology refinement.

**Figure 2. Ontology Builder pipeline in publication form.** The diagram combines the steps explicitly requested in the task formulation with a feedback loop "response → fact extraction → graph." Architecturally, this corresponds to modern work on LLM-supported ontology engineering and external-context systems.

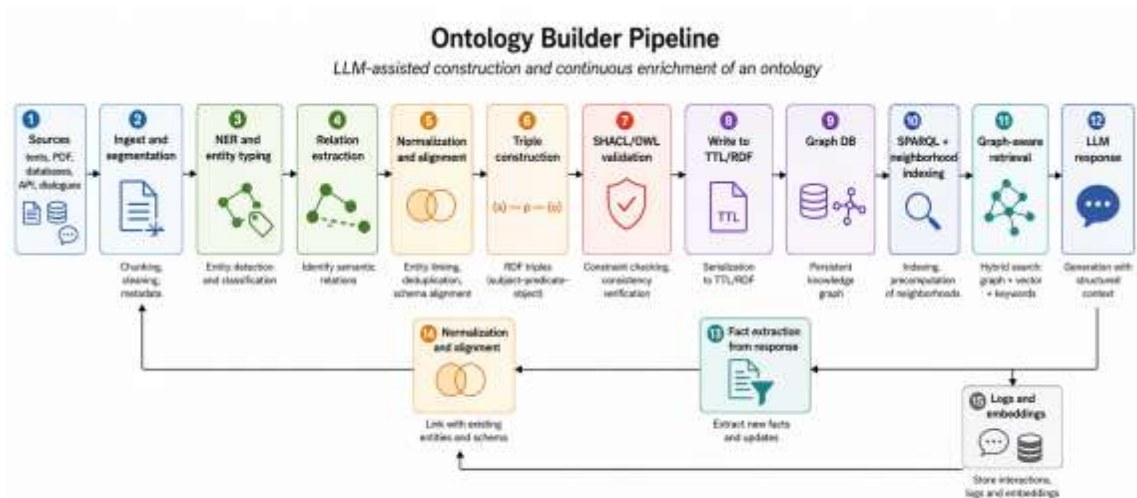

Figure 2. Ontology Builder pipeline in publication form. The diagram combines the steps explicitly requested in the task formulation with a feedback loop "response → fact extraction → graph." Architecturally, this corresponds to modern work on LLM-supported ontology engineering and external-context systems. [21]

```
flowchart LR
    S[Sources: texts, PDF, databases, API, dialogues] --> I[Ingest and segmentation]
    I --> NER[NER and entity typing]
    NER --> RE[Relation extraction]
    RE --> NORM[Normalization and alignment]
    NORM --> TRI[Triple construction]
    TRI --> VAL[SHACL/OWL validation]
    VAL --> TTL[Write to TTL/RDF]
    TTL --> KG[Graph DB]
    KG --> IDX[SPARQL + neighborhood indexing]
    IDX --> RET[Graph-aware retrieval]
    RET --> LLM[LLM response]
    LLM --> FACT[Fact extraction from response]
    FACT --> NORM
    LLM --> LOGS[Logs and embeddings]
    LOGS --> I
```

### Stage 1 — Ingest and Segmentation

Sources enter the system in heterogeneous form: documents, PDFs, tables, transcripts, API responses, and dialogues. For texts and logs, standard RAG-like chunking and embedding indexing are performed, providing inexpensive recall via similarity.

A P.Salovskii's perspective is useful here: RAG is not rejected but interpreted as connecting external memory, which remains useful as long as corpus recall solves part of the problem. However, already at this stage it is crucial to preserve provenance: which chunk, from which source, in which context produced a future entity or relation.

**Stage 2 — NER and Entity Typing**

The system extracts candidate entities, assigns preliminary types and aliases, and then converts detected entities into vocabulary candidates for the future ontology.

Ontology engineering literature emphasizes that most future errors originate at this step: incomplete classes, confusion between instance and type, ambiguous aliases, incorrect domain and range definitions. Therefore, the output of NER must be treated as a hypothesis space rather than a finalized model.

**Stage 3 — Relation Extraction**

Here, the LLM extracts not merely pairs of related entities but candidates for predicates and their possible semantics: *part-of, located-in, used-for, prohibited-by, depends-on, causes, precedes*, etc.

Work by Feng et al. shows that grounding in a target ontology or schema significantly improves interpretability and consistency of the extracted graph. Similarly, Aggarwal et al. demonstrate that modern models can reliably distinguish broader/narrower/same-as relationships between topics, meaning that relation induction can already be automated at a level sufficient for a productive ontology draft.

**Stage 4 — Normalization and Alignment**

Entities detected across different sources must be mapped to stable identifiers, synonyms consolidated, homonymy resolved, and duplicates merged.

At this stage, problems such as "meters vs yards," "obsolete class versions," "different formulations of the same process," or "the same object under different URIs" are addressed.

The supplied seminar notes explicitly identify this as the stage where the ontology must accumulate both historical and alternative interpretations of an entity. Modern ontology engineering surveys also emphasize that reuse, maintenance, and validation are not afterthoughts but central elements of the ontology lifecycle.

**Stage 5 — Triple Construction**

After normalization, the builder serializes facts into subject–predicate–object tuples. This makes knowledge machine-readable in the strict sense: relationships cease to be implicit in text and become first-class objects of the graph.

In the author's article series on Habr, this is described as the transition from "similar fragments" to "explicit relations," and in RDF specifications as the shift to a graph-based data model.

This is where the infrastructural advantage of ontologies appears: instead of storing textual wrappers, the system stores structures directly usable for graph traversal, filtering, reasoning, and explanation.

**Stage 6 — Validation**

Validation must operate in two modes:

- **Structural validation**, via SHACL shapes:
  which classes must have which properties, where cardinality constraints apply, which values are expected, and what typing is considered correct.

- **Logical validation**, via OWL reasoning:
  whether classes are compatible, whether contradictions have emerged, whether hierarchies are correct, and whether new inferences can be derived.

In practice, this is implemented using libraries such as Apache Jena SHACL for shape validation, or OWL reasoners like Pellet, capable of checking consistency, performing classification, and explaining inferences.

**Stage 7 — Writing to TTL/RDF and Publishing to Graph DB**

In the project service architecture, this step is explicitly defined as the transition:

**Ontology Builder code/service → TTL file → Graph DB**

This is important not only as a technical detail but as a scientifically reproducible artifact: TTL/RDF serves as an intermediate, serializable, versionable representation of knowledge.

Through it, one can perform CI-like validation, compare deltas, execute ontology diff operations, build provenance traces, and reproduce reasoning outside the main system.

The design of RIGOR and related works also relies on incremental ontology fragments, making this approach fully aligned with modern practice.

**Feedback Loop**

A closed feedback loop is then activated. Dialogue and system responses are not merely archived but are reprocessed through fact extraction.

If the LLM produces a new fact, judgment, relation, or state, it becomes part of the candidate ontology delta. If the fact fails validation, it remains in logs but does not enter the trusted graph.

This mechanism generates answers are shown separately from verdicts (supported/contradicted), indicating a two-loop system: generation plus ontological or regulatory verification.

At the next level, this loop naturally extends into the **agent layer**, where specialized agents handle retrieval, validation, planning, execution, and audit trails.

Theoretically, such decomposition aligns well with both context-centric architectures and modern arguments in favor of multi-model agentic systems.

Here is a **complete, accurate scientific translation**, preserving the **entire structure, formatting, tables, diagram code, and references** without reduction or reinterpretation:

# Experiments and Results

The empirical part of the paper is deliberately divided into quantitative and qualitative blocks. The quantitative block is based on the provided Tower of Hanoi benchmark chart; the qualitative block is based on Fact Analyzer screenshots.

Such a design is honest with respect to the source data: it does not pretend that we are dealing with a fully specified benchmark suite, but it also does not ignore important project observations. At the same time, the chosen quantitative task is not accidental. External literature on planning repeatedly shows that autoregressive LLMs have persistent limitations in planning tasks and reasoning about change, and benefit from tighter integration with external model-based verifiers. Therefore, the Tower of Hanoi represents a substantively appropriate testbed for evaluating the hypothesis of the usefulness of an ontological layer.

**Figure 3.** Reconstruction of the Tower of Hanoi diagram based on the provided image. The values were manually reproduced from the project chart; protocol details, sample size, and confidence intervals are unspecified in the supplied materials.

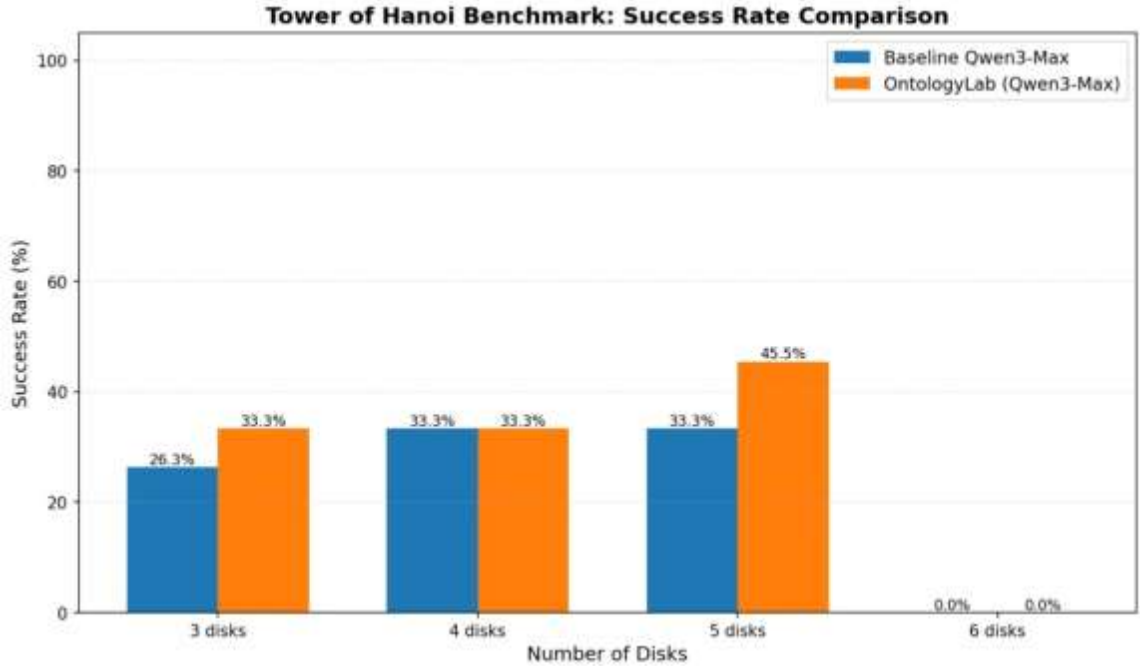

Below are the numerical values reconstructed from the supplied chart.

| Number of disks | Baseline Qwen3-Max | Ontology-augmented Qwen3-Max | Absolute change |
|---|---|---|---|

| Number of disks | Baseline Qwen3-Max | Ontology-augmented Qwen3-Max | Absolute change |
|---|---|---|---|
| 3 | 26.3% | 33.3% | +7.0 p.p. |
| 4 | 33.3% | 33.3% | 0.0 p.p. |
| 5 | 33.3% | 45.5% | +12.2 p.p. |
| 6 | 0.0% | 0.0% | 0.0 p.p. |

Even in this limited form, the pattern is indicative. For 3 disks, a moderate improvement is observed; for 4 disks, no difference; for 5 disks, the most significant gain; and for 6 disks, both configurations fail. The most cautious interpretation is that the ontological layer begins to provide benefit when the task already exceeds trivial local search but has not yet become entirely intractable for the chosen model configuration. In other words, symbolic structure and verifiable constraints are especially useful in the "middle complexity zone," where an LLM without an external world model has not yet completely collapsed but already systematically loses plan fidelity. This interpretation is the author's conclusion based on the supplied chart and the general literature on the difficulties of planning using LLMs alone.

The qualitative block supports the same hypothesis from a different perspective.

Figure 4. Fact Analyser screenshot

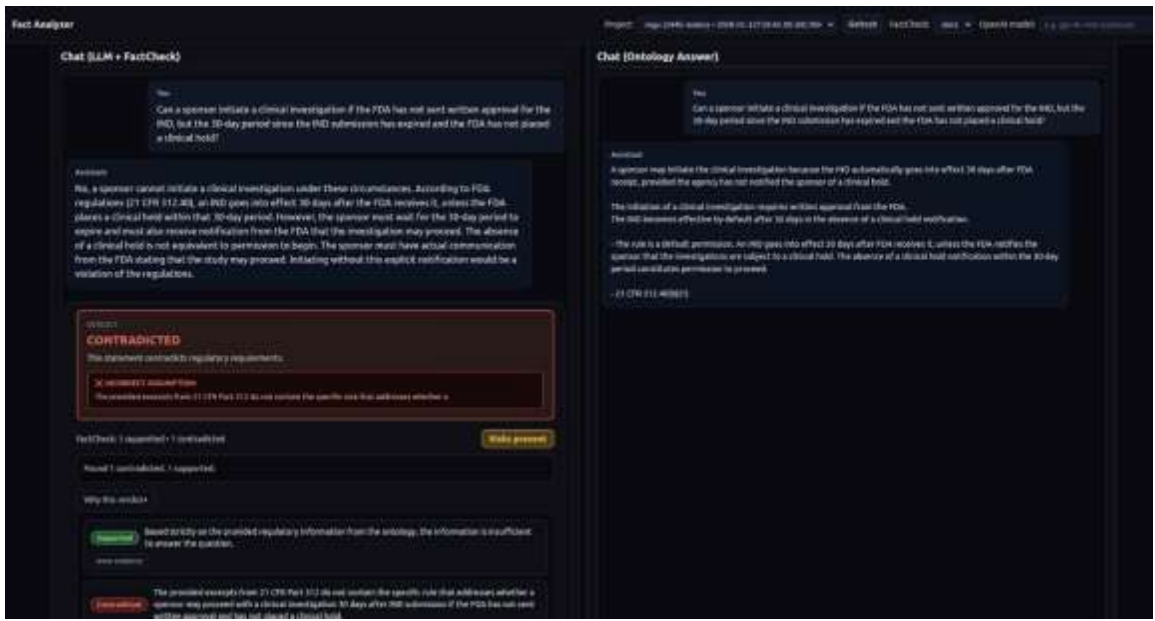

In the supplied Fact Analyzer screenshot, a regulatory question is presented regarding the possibility of initiating a clinical investigation after the expiration of a 30-day period following IND submission, in the absence of written approval and in the absence of a clinical hold. In the left panel, the "LLM + FactCheck" system generates a negative answer and receives a strict verdict of **CONTRADICTED**. In the right panel, the "Ontology Answer" produces a structured response, according to which the sponsor may initiate the investigation, since the IND becomes effective after 30 days in the absence of a clinical hold notification; this response is visually aligned with an ontology-backed citation fragment. Even without access to the full dataset, the screenshot demonstrates an important point: when formalized knowledge is present, verification is performed not based on general "plausibility" of the text, but on a specific rule and its conditions of applicability.

From an engineering perspective, this means that the actual advantage of ontology lies not only in improved retrieval but also in the emergence of a separate verification channel. A pure LLM generates an answer. The ontology and validator check it against a trusted representation of the rule, and may then either accept it, reject it, or annotate it with a trace. It is precisely this two-loop scheme that this work proposes to interpret as explainable neuro-symbolic QA.

At the same time, both empirical parts share a common limitation. For the Tower of Hanoi, the number of episodes, aggregation method, prompt templates, stop conditions, and success criteria are unknown. For the Fact Analyzer, the composition of the corpus, coverage of regulatory norms, conflict resolution policy, and final verdict labeling criteria are unknown. In all cases, these parameters should be considered unspecified. Therefore, the results should be used appropriately as proof-of-concept and architectural evidence, but not as a finalized statistical comparison with external benchmark papers.

## Discussion and Limitations

The main conclusion from literature and the supplied materials is that ontology enriches retrieval along three distinct axes. The first is **structural**: a query can be directed not only to embedding neighborhood, but also to graph neighborhood, types, paths, constraints, and causal dependencies. The second is **validation**: before a new fact becomes part of trusted knowledge, it passes SHACL/OWL checks. The third is **memory**: the system accumulates not a stream of text, but a mutable model of the world, the user, and the process. It is precisely this that shifts the system from the mode of "retrieving another piece of context" to the mode of "updating the state of an interpretable model."

For agent-based systems, this is particularly significant. It is not enough for an agent to generate a well-formed text; it requires a world in which entities, predicates, action predicates, permissible transitions, constraints, and consequences are defined. In such a setting, planning can be organized not as purely linguistic improvisation, but as a cycle: **proposal → symbolic check → repair → execution**. It is exactly this tighter bi-directional interaction between LLM and external model-based verifiers that is recommended by Kambhampati et al.; context-centric systems such as Interfaze, as well as the economic justification of SLM-rich agentic stacks presented by Belcak et al., make this conclusion practically realistic. In such a configuration, the Ontology Builder and the reasoning layer play the role of a "slow but reliable" core, while specialized agents and small models serve as inexpensive operational modules around it.

From the perspective of the path toward stronger forms of AI, new opportunities emerge. First, an ontological long-term memory appears, suitable for multi-session interaction, personalization, and robotic state tracking. Second, explainable reasoning becomes possible: the path from an answer to facts and rules can be traced. Third, conditions arise for stateful agents that not only read documents but also operate on artifacts of the world— contracts, objects, process states, technological constraints, dialogue history, and actions of other agents. Fourth, data integration improves: through ontology, documents, APIs, logs, data lakes, and tabular systems can be unified within a single semantic layer. Such scenarios are directly consistent both with the publicly available description of the AGI seminar and with recent academic literature on ontology engineering, GraphRAG, and agentic systems.

However, this program has substantial limitations. First, automatic ontology construction is still not equivalent to error-free ontology engineering: surveys and benchmarking studies emphasize variability, schema drift, hallucinated relations, reuse problems, and the need for multi-faceted quality evaluation beyond human-like plausibility. Second, the pipeline requires continuous normalization and alignment; otherwise, the graph quickly turns into "sophisticated chaos," as P.Salovskii describes this risk. Third, reasoning and validation introduce latency and require careful orchestration: not every query should trigger a full reasoning pass. Fourth, reducing manual labor does not imply its elimination; the literature consistently recommends human-in-the-loop at least at the level of schema governance, evaluation criteria, and release management.

The limitations of the present article are compounded by the limitations of the corpus. Since part of the supplied materials is provided only as images rather than reproducible experimental packages, the results section is necessarily descriptive in nature.

## Discussion and Conclusion

The conducted analysis allows the final thesis to be formulated in максимально сжатом виде: an LLM by itself remains a language engine, whereas ontology transforms the system into a machine for working with knowledge. This transformation occurs not because a graph is "better than text in general," but because a graph introduces explicit entities, relationships, constraints, validation procedures, and long-term memory where a pure LLM possesses only probabilistic generation and short-lived context. The combined use of MCP, RDF/OWL, SPARQL, SHACL, reasoning engines, vector RAG, and an agent layer forms an architecture in which the functions of generation, memory, retrieval, and verification are separated and therefore can reinforce each other rather than compete.

From a practical perspective, the supplied materials confirm precisely this architectural shift. The Tower of Hanoi demonstrates that an ontological layer can improve success rates in planning tasks, at least within a certain range of complexity. The Fact Analyzer shows that the combination of "generation + formal verification" can distinguish a supported answer from a contradicted answer. The P.Salovskii's article series convincingly establishes the philosophical and engineering framework: the transition from text to knowledge and from dialogue history to ontological memory. Together, these materials do not constitute a complete body of proof, but rather form a research program.

Logical directions for future work include: the publication of a reproducible benchmark suite for ontology-augmented planning; the introduction of temporal and versioned ontologies to support the historicity of norms and states; automatic management of measurement units and alternative vocabularies; probabilistic confidence layers over triples; integration of graph memory with tool-using agents and robotic action models.